\documentclass{article}
\usepackage[utf8]{inputenc}
\usepackage{amsfonts}
\usepackage{amsmath}
\usepackage{xcolor}
\usepackage{iclr2021,times}
\usepackage{graphicx}
\usepackage{float}
\usepackage{subcaption}
\usepackage{algorithm}
\usepackage{algpseudocode}
\usepackage{algorithmicx}

\usepackage{amsmath,amsfonts,bm}

\def\eqref#1{equation~\ref{#1}}

\def\1{\bm{1}}

\DeclareMathAlphabet{\mathsfit}{\encodingdefault}{\sfdefault}{m}{sl}
\SetMathAlphabet{\mathsfit}{bold}{\encodingdefault}{\sfdefault}{bx}{n}

\usepackage{hyperref}
\usepackage{url}

\title{Leveraging the Variance of Return Sequences for Exploration Policy} 

\author{Zerong Xi\\
Department of Computer Science\\
University of Central Florida\\
Orlando, FL 32765, USA \\
\texttt{xizerong@gmail.com} \\
\And
Gita Sukthankar\\
Department of Computer Science\\
University of Central Florida\\
Orlando, FL 32765, USA \\
\texttt{gitars@eecs.ucf.edu} \\
}

\begin{document}

\maketitle

\begin{abstract}
This paper introduces a method for constructing an upper bound for exploration policy using either the weighted variance of return sequences or the weighted temporal difference (TD) error.  We demonstrate that the variance of the return sequence for a specific state-action pair is an important information source that can be leveraged to guide exploration in reinforcement learning.  The intuition is that fluctuation in the return sequence indicates greater uncertainty in the near future returns.  This  divergence occurs because of the cyclic nature of value-based reinforcement learning; the evolving value function begets policy improvements which in turn modify the value function.  Although both variance and TD errors capture different aspects of this uncertainty, our analysis shows that both can be valuable to guide exploration.  We propose a two-stream network architecture to estimate weighted variance/TD errors within DQN agents for our exploration method and show that it outperforms the baseline on a wide range of Atari games.
\end{abstract}

\section{Introduction}
Having a good exploration policy is an essential component of achieving sample efficient reinforcement learning.   Most RL applications use two heuristics, visitation counts and time, to guide exploration.  \textit{Count-based exploration}~\citep{uct} assumes that it is worth allocating the exploration budget towards previously unexplored actions by awarding exploration bonuses based on action counts. \textit{Time-based exploration}~\citep{RLsurvey} is usually implemented using a Boltzmann distribution that reduces exploration during later stages of the learning process.  This paper presents an analysis of the benefits and drawbacks of weighted sequence variance for guiding exploration; we contrast the performance of weighted variance with the more familiar weighted temporal difference (TD) error.

Our intuition about the merits of weighted variance as a heuristic to guide exploration is as follows. Imagine that the returns are being summed in a potentially infinite series. Weighted variance can be computed online in order to estimate the convergence speed of the series for a specific state-action pair.  We construct an upper bound using uncertainty, modeled as the weighted standard deviation, as an exploration bonus to guide action selection. Fluctuation in the return sequence may foretell greater uncertainty in the near future returns that should be rectified through allocation of the exploration budget.  Value-based RL algorithms are particularly susceptible to divergence, since improvements in the value function result in rapid policy changes which in turn affect the value estimation. Unlike event counts, weighted variance is more sensitive to the dynamics of the return sequence; if multiple visitations yield consistent reward, weighted variance will quickly prioritize a different state-action sequence even if the total event counts are smaller. 

However, computing weighted variance within a deep reinforcement learning framework is a challenging problem, due to the instability of deep neural networks.  Simply computing the variance directly from output of DQN risks overestimating the error. The second contribution of the paper is introducing two-stream network architecture to estimate either weighted variance or TD errors within DQN agents.  Our new architecture (Variance Deep Q Networks) uses a separate $\sigma$ stream to estimate a weighted standard deviation of the outputs from the original stream.


\section{Related Work}

Several groups have proposed strategies for balancing exploration/exploitation in deep reinforcement learning including 1) extensions on count-based methods~\citep{Bellemare2016unifying,Ostrovski2017count, Tang2017exploration}; 2) noise injection techniques~\citep{Fortunato2018noisy, Plappert2018parameter}; 3) improving uncertainty estimation~\citep{Osband2016bootstrap,Nikolov2018informationdirected}; 4) driving exploration with intrinsic motivation~\citep{Saul2005intrinsically,Pathak2017curiosity} and 5) entropy-guided architectures~\citep{Haarnoja2017energy,Hazan2019entropy}.  Our proposed weighted-variance guided exploration technique is a compatible addition to some of these other techniques (see Section~\ref{section:discussion} for further details).

Moving from tabular to deep reinforcement learning makes the problem of estimating quantities such as counts and variance more challenging.  \cite{Ostrovski2017count} showed how count-based techniques could be generalized to neural networks by learning a density model on state sequences; pseudocounts for states are then derived from the density model's recoding probability. In contrast our model learns the weighted variance over the return sequence rather than the state sequence. 

Bootstrapped DQN~\citep{Osband2016bootstrap} uses the classic bootstrap principle to estimate uncertainty over Q-values.   Agent actions are selected using a single Q function sampled from the posterior distribution.  Rather than dithering like noise based models, bootstrapped DQN promotes deep exploration by maintaining policy continuity with regards to the single sampled Q-function.  Uncertainty in our proposed architecture is quantified by the weighted standard deviation of the return sequences instead of through data partitions, but like Bootstrapped DQN, our architecture uses multiple (two) heads. 

The use of randomness or noise to drive exploration is a common theme across many approaches.  NoisyNets~\citep{Fortunato2018noisy} directly injects noise into the weights of the neural networks; the parameters of the noise are learned in combination with the network weights. Like NoisyNet, our uncertainty is learned directly by the network, reducing the need for extra hyperparameters.

Prior work has also analyzed the dynamics of value functions in reinforcement learning.
\cite{Sutton2018rl} show how policy and value functions change throughout the learning process.
Like them, we believe that the variation of the sequence is predictive of the uncertainty of the near future returns. Variation in the return sequence can be modeled as a slower convergence trend layered with transient fluctuations. Our method builds on this finding; see Appendix~\ref{appendix:empirical_analysis} for an empirical analysis showing how weighted variance reacts to the dynamics of raw, smoothed, and residual return sequences.





\section{Background}
Our aim is to learn an action policy for a stochastic world modeled by a Markov Decision Process by balancing the exploration of new actions and the exploitation of actions known to have a high reward.  This is done by learning a value function ($Q(s,a)$) using the discounted return information ($G(s,a)$) and learning rate ($\alpha$):  


\begin{equation}
    Q(s,a) = Q(s,a) + \alpha \cdot (G(s,a) - Q(s,a))
    \label{eq:q_update}
\end{equation}

Actions are selected using the learned value function.  This paper illustrates how our weighted variance exploration approach can be integrated into agents using deep Q-learning.


\textbf{Deep Q Networks}~\citep{Mnih2015dqn} utilize deep neural networks as approximators for action-value functions in Q learning algorithms. The updating procedure of the function is formulated as an optimization problem on a loss function:
\begin{equation}
    L_{\text{DQN}} (\zeta) = \mathbb{E}_{(s, a, r, s') \sim D} \left[ \left(r + \gamma \cdot \text{max}_{b \in \mathcal{A}} Q(s', b; \zeta^-) - Q(s, a; \zeta) \right)^2 \right]
    \label{eq:dqn_loss}
\end{equation}
where $\zeta$ are the parameters of the network, $\mathcal{A}$ is a set of valid actions and $D$ is a distribution  over a replay buffer of previously observed transitions. A target network with parameters $\zeta^-$ is regularly synchronized with $\zeta$ and used to estimate the action values of the successor states; the use of a target network promotes estimation stability.  Since the original introduction of DQN, several improvements to the updating procedure and network architecture have been proposed.

\textbf{Double DQN}~\citep{Hasselt2016double} updates the network according to a different rule in which the action for the successor state is selected based on the target network rather than the updating network. This change alleviates the overestimation problem by disentangling the estimation and selection of action  during optimization steps. The loss function for Double DQN is:
\begin{equation}
    L_{\text{DDQN}} = \mathbb{E}_{(s, a, r, s') \sim D} \left[ \left(r + \gamma \cdot Q(s', \text{argmax}_{b \in \mathcal{A}} Q(s', b; \zeta); \zeta^-) - Q(s, a; \zeta) \right)^2 \right].
    \label{eq:ddqn_loss}
\end{equation}

\textbf{Dueling DQN}~\citep{Wang2016duel} uses a two-stream network architecture to separate the estimation of state value and action advantage in order to accelerate learning in  environments with a large number of valid actions. The estimation of Q value is a function of value stream $V(\cdot, \cdot; \cdot)$ and advantage stream $A(\cdot, \cdot; \cdot)$ such that
\begin{equation}
    Q(s, a; \zeta) = V(s, a; \zeta) + A(s, a; \zeta) - \frac{\sum_{b \in \mathcal{A}} A(s, b; \zeta)}{N_{\text{actions}}}.
\end{equation}

\textbf{Prioritized Replay}~\citep{Schaul2015per} selects the experience used for the optimation procedure from a replay buffer with a priority-based sampling method instead of a uniform one.  Sampling probabilities are  proportional to the TD errors.

Our proposed technique guides exploration and can be coupled with other improvements to the reinforcement learning process. 
Although we construct our models using the above extensions, our technique is solely for guiding exploration and can benefit from other improvements, such as Bootstrapped DQN~\citep{Osband2016bootstrap}, Distributional DQN~\citep{Bellemare2017distributional,Dabney2018distributional}, and Multi-Step Learning~\citep{Sutton2018rl}. Moreover, it can be combined with the other exploration methods, such as count-based methods~\citep{Ostrovski2017count} or Noisy DQN~\citep{Fortunato2018noisy}, to guide exploration during different training stages. We discuss this further in Section~\ref{section:discussion}.

\section{Measuring Variance for Exploration}

During Monte Carlo policy evaluation, the value function $Q(s,a)$ for a particular state-action pair is updated using a sequence of returns $\mathcal{G}_n(s,a) = G_1(s,a), G_2(s,a),...,G_n(s,a)$.
This series can start diverging due to the cyclic nature of value-based approaches; the evolving value function begets policy improvements which in turn modify the value function.
We believe that the agent should leverage information from these variations to quantify uncertainty in order to explore non-optimal, but still promising, actions.
Specifically, agents can follow a greedy exploration policy based on an upper bound:
\begin{equation}
    \pi(s) = \text{argmax}_{a \in \mathcal{A}} Q(s,a) + \sigma (s,a) \cdot c ,
    \label{eq:upper_bound}
\end{equation}
where $\sigma$ is a measurement of uncertainty and $c$ is a fixed hyper-parameter which adjusts the extent of exploration.

To measure the uncertainty of returns for a specific state-action pair, we propose 1) a weighted variance estimation method for general RL, 2) a neural network architecture, and 3) novel optimizing strategy which explicitly estimates either weighted variance or weighted TD error in the DRL configuration.

\subsection{Reinforcement Learning with Variance Estimation}

Although the vanilla form of sequence variance doesn't reflect the higher importance of the recent returns, we define the uncertainty as an exponentially decaying weighted standard deviation
\begin{equation}
    \label{eq:ve_variance}
    \sigma_n(s,a) = \sqrt{ \frac{\sum_{i=1}^n(1-\alpha)^{n-i}(G_i(s,a) - Q_n(s,a))^2}{\sum_{i=1}^n(1-\alpha)^{n-i}} },
\end{equation}
where $Q_n(s,a)$ is the value function which is updated using $\mathcal{G}_n(s,a)$ and $\alpha$ (the update step size) in equation~\ref{eq:q_update}.

The update formula for $\sigma$ is as follows
\begin{equation}
    \label{eq:ve_update}
    \sigma_{n+1} (s,a) = \sqrt{(1 - \alpha) \left[ \sigma_n^2 (s,a) + (Q_{n+1} (s,a) - Q_{n} (s,a))^2 \right] + \alpha (G_{n+1} (s,a) - Q_{n+1} (s,a))^2}.
\end{equation}
The first term inside the square root represents the adjusted estimation of variance on $\mathcal{G}_n(s,a)$ with the updated $Q_{n+1}(s,a)$, and the second term is the estimation from the incoming $G_{n+1}(s,a)$.
The details of the updating formula are presented in Appendix~\ref{appendix:updating_formula}.

When updates are performed using the above formula, $\sigma(s,a)$ is biased during the early stage, due to the undecidable prior $\sigma_0(s,a)$ as well as the bias incipient to the usage of a small $n$. We propose two strategies for initializing the $\sigma$ function: 1) warming up with an $\epsilon$-greedy method with $\epsilon$ decayed to 0 to ensure a gradual transition to our exploration policy, which effectively starts with a larger $n$; 2) initializing $\sigma_0(s,a)$ as a large positive value to encourage uniform exploration during early stages, which is theoretically sound since the variance of the value of a state-action pair is infinitely large if it has never been visited.

\subsection{Variance Deep Q Networks (V-DQN)}

Our new algorithm, V-DQN, incorporates weighted variance into the exploration process of training DQNs.

\begin{algorithm}
\caption{Variance DQN (V-DQN)}
\hspace*{\algorithmicindent} \textbf{Input:} exploration parameter $c$; minibatch $k$; target network update step $\tau$; \\
\hspace*{\algorithmicindent} \textbf{Input:} initial network parameters $\zeta$; initial target network parameter $\zeta^-$; \\
\hspace*{\algorithmicindent} \textbf{Input:} Boolean DOUBLE
\begin{algorithmic}[1]
\State Initialize replay memory $\mathcal{H} = \emptyset$
\State Observe $s_0$
\For {$t \in \{1,...,T\}$}
    \State Select an action $a \gets$argmax$_{b \in \mathcal{A}} Q(s, b; \zeta) + |\sigma(s, b; \zeta)| \cdot c$
    \State Sample next state $s \sim P(\cdot|s, a)$ and receive reward $r \gets R(s, a)$
    \State Store transition $(s_{t-1}, a, r, s_t)$ in $\mathcal{H}$
    \For {$j \in \{1,...,k\}$}
        \State Sample a transition $(s_j, a_j, r_j, s'_j) \sim D(\mathcal{H})$  \Comment{$D$ can be uniform or prioritised replay}
        \If {$s'_j$ is a terminal state}
            \State $G \gets r_j$
        \ElsIf {DOUBLE}
            \State $b^\ast(s'_j) = \text{argmax}_{b \in \mathcal{A}} Q(s'_j, b; \zeta)$
            \State $G \gets r_j + \gamma Q(s'_j, b^\ast(s'_j);\zeta^-)$
        \Else
            \State $G \gets r_j + \gamma \text{max}_{b \in \mathcal{A}} Q(s'_j, b; \zeta^-)$
        \EndIf
        \State $\hat \sigma \gets G - Q(s_j, a; \zeta)$
        \State Do a gradient step with loss $(G - Q(s_j, a; \zeta))^2 + ({\hat \sigma}^2 - \sigma^2(s_j, a; \zeta))^2$
    \EndFor
    \If {$t \equiv 0\;(\text{mod}\;\tau)$}
        \State Update the target network $\zeta^- \gets \zeta$
    \EndIf
\EndFor
\end{algorithmic}
\label{algo:vdqn}
\end{algorithm}

Due to the known instability of deep neural networks during the training process, it is risky to calculate the weighted variance from composing multiple estimations (e.g., the state-action values before and after the optimization step).
Instead of computing the target variance as a byproduct, we propose a two-stream neural network architecture along with an novel loss function to allow end-to-end training while estimating the weighted standard deviation.
It simplifies the optimization for variance by sacrificing the accuracy, which we believe is an acceptable compromise since the deep neural networks with gradient descent cannot strictly follow the above updating formula. Our empirical results demonstrate the effectiveness of the simplification.

Variance DQN uses neural networks with a separate $\sigma$-stream to estimate a weighted standard deviation of the outputs from the original stream on moving targets, which is common in the context of deep reinforcement learning where the value function improves as the policy evolves. In practice, the $\sigma$-stream shares lower layers, e.g. convolutional layers, with the original stream to reduce computational demands.

The loss function for Variance DQN is a sum of mean square error losses on the original stream, which is identical to formula~\ref{eq:dqn_loss} (for DQN) or formula~\ref{eq:ddqn_loss} (for Double DQN), and the square of the $\sigma$-stream:
\begin{align}
    & L_{\text{V-DQN}} = \mathbb{E}_{(s, a, r, s') \sim D} \left[ \left( G - Q(s, a; \zeta) \right)^2 + \left( (G - Q(s, a; \zeta))^2 - \sigma^2(s, a; \zeta) \right)^2 \right] \\
    s.t. \;\;\; & G = 
    \begin{cases}
    \label{eq:g_value}
    r + \gamma \cdot \text{max}_{b \in \mathcal{A}} Q(s', b; \zeta^-) & \text{for DQN} \\
    r + \gamma \cdot Q(s', \text{argmax}_{b \in \mathcal{A}} Q(s', b; \zeta); \zeta^-) & \text{for Double DQN}
    \end{cases}
\end{align}
It is worth noting that the Q function used in the second part of the loss function on the $\sigma$-stream doesn't contribute to gradients directly. Therefore, the optimization steps are in effect unchanged for the original stream on  the Q value, except for the shared lower layers.  While the sign of the $\sigma$-stream's output is eliminated in the loss function, we need to do the same during the exploration process. The modified exploration policy is
\begin{equation}
    \pi (s) = \text{argmax}_{a \in \mathcal{A}} Q(s, a) + |\sigma(s, a)| \cdot c
\end{equation}
The full procedure is shown in Algorithm~\ref{algo:vdqn}.

We also propose a variant of our method (TD-DQN) which updates $\sigma$-stream with absolute temporal difference error. The loss function for TD-DQN is
\begin{equation}
    L_{\text{TD-DQN}} = \mathbb{E}_{(s, a, r, s') \sim D} \left[ \left( G - Q(s, a; \zeta) \right)^2 + \left( |G - Q(s, a; \zeta)| - \sigma(s, a; \zeta) \right)^2 \right]
\end{equation}
where $G$ is the same as Equation~\ref{eq:g_value}.

Both networks measure the uncertainty of the Q value based on the return history in order to construct an upper bound for exploration policy. The difference between the approaches can be interpreted based on their implicit usage of different distance metrics: Euclidean (Variance DQN) vs.\ Manhattan (TD-DQN). Generally, V-DQN is more sensitive to fluctuations in the return sequence, investing a greater amount of the exploration budget to damp variations.

There may be applications in which it is is valuable to estimate still higher order statistics, such as the skew or kurtosis of the return sequence.
However, this sensitivity can also sabotage exploration by directing resources away from promising areas of the state space that are slowly trending towards convergence;  overemphasizing the elimination of small variations could ultimately result in longer training times.
While the $c$ hyper-parameter in our exploration policy adjusts the trade-off between exploration and exploitation, the choice of the distance metrics used to measure sequence variation determines the distribution of exploration time.

\section{Results}


To illustrate how weighted variance improves exploration, this paper first presents results on its usage in tabular Q-learning for the Cartpole inverted pendulum problem.  Then we report the performance of our proposed techniques (V-DQN and TD-DQN) on the Atari game benchmark. The results show that our two stream architecture for guiding exploration with weighted variance or weighted temporal difference outperforms the standard DDQN benchmark.

\subsection{Cartpole}

To demonstrate the effectiveness of our Variance Estimation (VE) method, we compare it with $\epsilon$-greedy on Cartpole balancing problem.
For this experiment, we use the classic Q-learning algorithm~\citep{Watkins1992q-learning} with a tabular look-up Q-table.

The Cartpole environment has a 4-dimensional continuous state space $\mathcal{S}=\mathbb{R}^4$ and a discrete action space $\mathcal{A}=$\{\textit{Push Left, Push Right, Do Nothing}\}. It provides a reward of 1 in every time step.  The episode terminates if any of the following conditions is met: 1) the episode length is greater than 500, 2) the pole falls below a threshold angle, 3) the cart drifts from the center beyond a threshold distance. With this setting, the maximum accumulated reward any policy can achieve is 500. To apply the tabular Q-learning configuration, the continuous state space is discretized into 18,432 discrete states by dividing \{\textit{Cart Position, Cart Velocity, Pole Angle, Pole Velocity}\} into \{$12, 8, 16, 12$\} intervals.

\begin{figure}[H]
    \centering
    \begin{minipage}{.48\textwidth}
        \includegraphics[width=\textwidth]{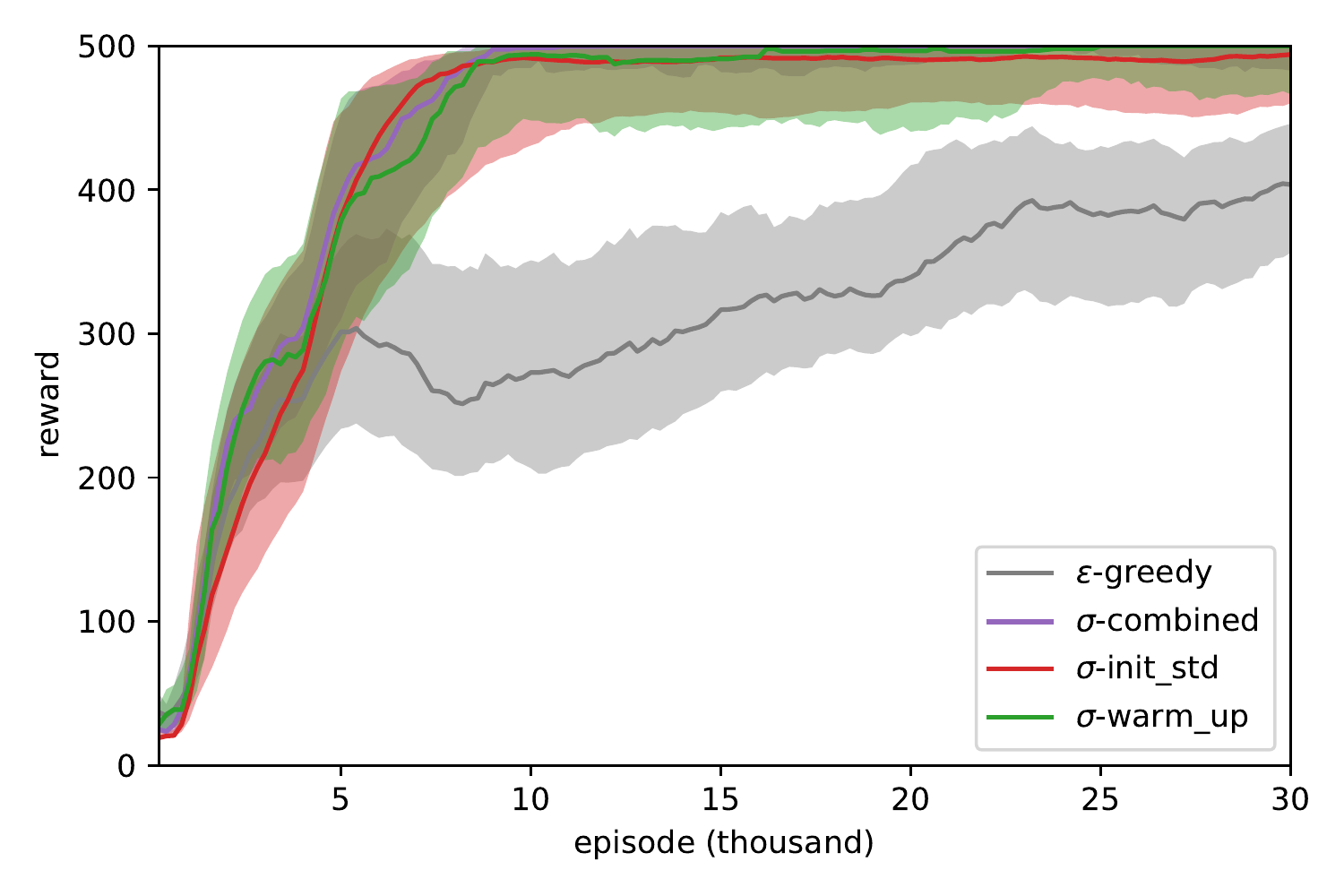}
        \subcaption{Episode reward ($\epsilon_{\text{eval}}=0$)}
    \end{minipage}
    \hfill
    \begin{minipage}{.48\textwidth}
        \includegraphics[width=\textwidth]{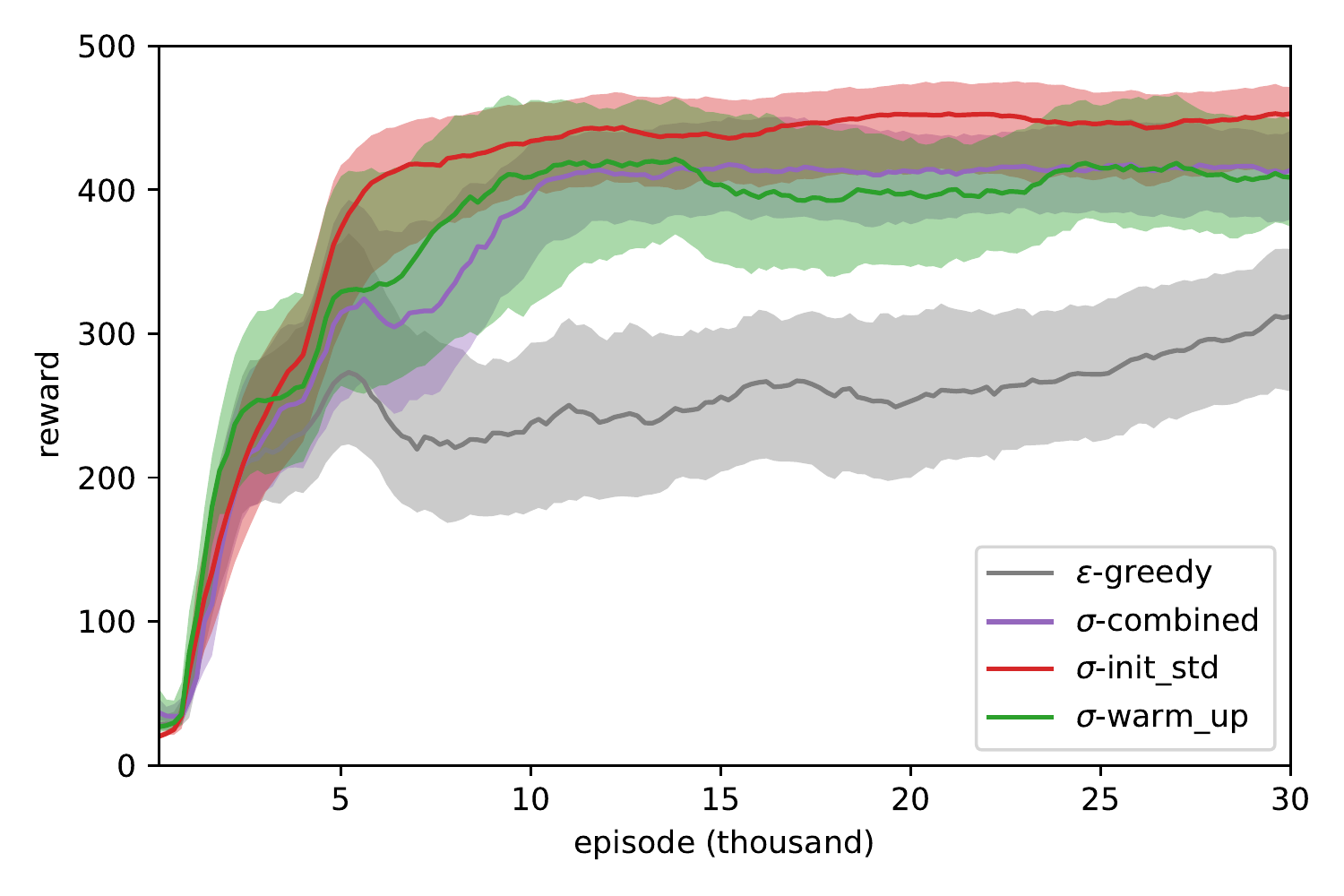}
        \subcaption{Episode reward ($\epsilon_{\text{eval}}=0.05$)}
    \end{minipage}
    \caption{Average episode rewards in the Cartpole balancing problem. The curves and the shadowed areas represent the means and the quartiles over 9 independent runs. The models are evaluated for 10 evaluation episodes every 200 training episodes.}
    \label{fig:mountain_car}
\end{figure}

With grid search, $\epsilon$-greedy achieves the best performance when the discounting factor $\gamma=1.0$ and the exploration rate $\epsilon$ decays from 1.0 to 0.01 in 5000 episodes.

For Variance Estimation, we experimented on both initialization methods as well as a combination of them. A similar configuration is applied on warming up with the $\epsilon$-greedy method in which $\epsilon$ decays from 1.0 to 0.0 during 5000 episodes. The initial standard deviation is set to 5000 for initializing the $\sigma_0$ method to ensure sufficient early visits on states. The combination method warms up with $\epsilon$-greedy while retaining the large initial standard deviation; it uses the same hyper-parameters. The value of $c$ is set to 1.5 for initializing the $\sigma_0$ method and 0.5 for the other two.

To reduce the possibility of over-fitting, we evaluate the models with an additional environment in which the agent has a probability $\epsilon_{\text{eval}}=0.05$ of acting randomly. All of our methods outperform $\epsilon$-greedy consistently for both evaluation settings. When the training time is prolonged, the baseline method generally achieves similar scores to our methods, but requires approximately 10 times the training episodes.


\subsection{Atari Games}

We evaluate our DQN-based algorithms on 55 Atari games from the Arcade Learning Environment (ALE)~\citep{Bellemare2013Ale}, simulated via the OpenAI Gym platform~\citep{OpenAI2016Gym}.
Defender and Surround are excluded because they are unavailable in this platform.
The baseline method (denoted as DDQN) is DQN~\citep{Mnih2015dqn} with all the standard improvements including Double DQN~\citep{Hasselt2016double}, Dueling DQN~\citep{Wang2016duel} and Prioritized Replay~\citep{Schaul2015per}.

Our network architecture has a similar structure to Dueling DQN~\citep{Wang2016duel}, but with an additional $\sigma$-stream among fully connected layers.
The 3 convolutional layers have 32 8$\times$8 filters with stride 4, 64 4$\times$4 filters with stride 2, 64 3$\times$3 filters with stride 1 respectively.
Then the network splits into three streams of fully connected layers, which are value, advantage and $\sigma$ streams.
Each of the streams has a hidden fully connected layer with 512 units.
For the output layers, the value stream has a single output while both advantage and $\sigma$ streams have the same number of outputs as the valid actions.

The random start no-op scheme in~\cite{Mnih2015dqn} is used here in both training and evaluation episodes. The agent repeats no-op actions for a randomly selected number of times between 1 to 30 in the beginning to provide diverse starting conditions to alleviate over-fitting. Evaluation takes place after freezing the network every 250K training steps (1M frames). The scores are the averages of episode rewards over 125K steps (500K frames) where episodes are truncated at 27K steps (108K frames or 30 minutes of simulated play).

We use the Adam optimizer~\citep{Kingma2015adam} with a learning rate of $6.25 \times 10^{-5}$ and a value of $1.5 \times 10^{-4}$ for Adam's $\epsilon$ hyper-parameter over all experiments. The network is optimized on a mini-batch of 32 samples over prioritized replay buffer every 4 training steps. The target network is updated every 30K steps.

The exploration rate of DDQN decays from 1.0 to 0.01 in 250K steps (1M frames) and retains that value until the training ends. Our methods do not rely on $\epsilon$-greedy so that is simply set to 0 for all the steps. Instead, the value of $c$ impacts the actual exploration rates of our methods, which are defined here to be the proportion of actions different from the optimal ones based on the current Q value function. Empirically, the performance on Atari games does not vary significantly over a wide range of $c$ values, which is an unusual finding. A possible explanation is that most of the actions in those games are not critical. To keep the exploration policy from drifting too far from the exploitation policy, we set $c$ to be 0.1 for both V-DQN and TD-DQN over all experiments to keep the average exploration rates of the majority of the Atari games to reside roughly between 0.01 and 0.1.

A summary of the results over all 55 Atari games is reported in Table~\ref{tab:atari_summary}. The raw and the normalized scores for individual games are compiled in Table~\ref{tab:atari_raw} and Table~\ref{tab:atari_normalized} in Appendix~\ref{appendix:results}.
To compare the performance of agents over different games, the scores are normalized with human scores
\begin{equation}
    \text{Score}_{\text{Normalized}} = 100 \times \frac{\text{Score}_{\text{Agent}} - \text{Score}_{\text{Random}}}{\text{Score}_{\text{Human}} - \text{Score}_{\text{Random}}}
\end{equation}
where both the random and the human scores were taken from \cite{Schaul2015per}. 

\begin{figure}
    \centering
    \includegraphics[width=\textwidth]{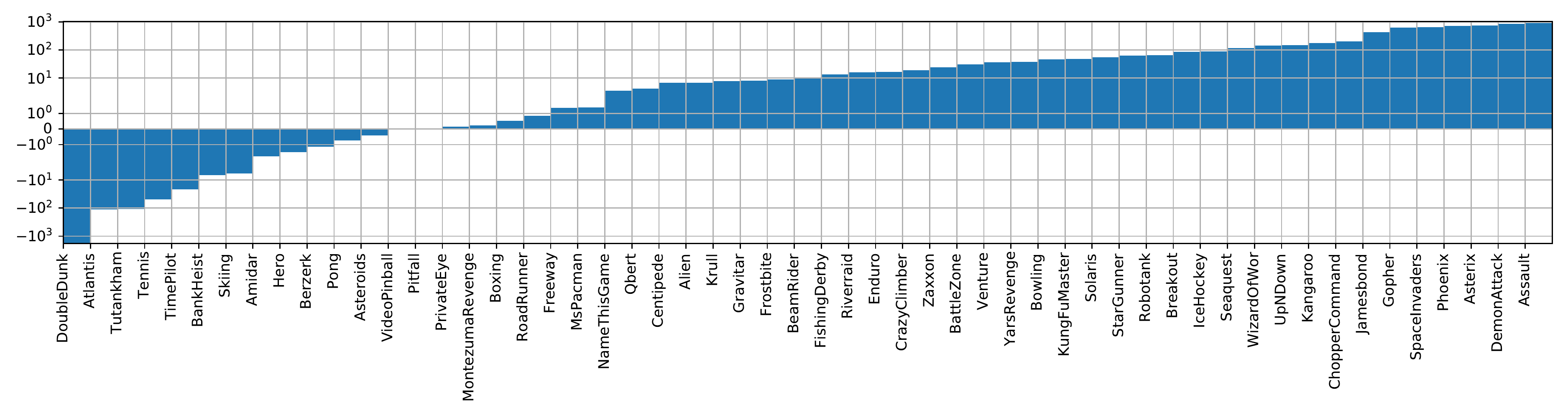}
    \caption{Improvement in normalized scores of V-DQN over DDQN in 200M frames}
    \vspace{0.5cm}
    \includegraphics[width=\textwidth]{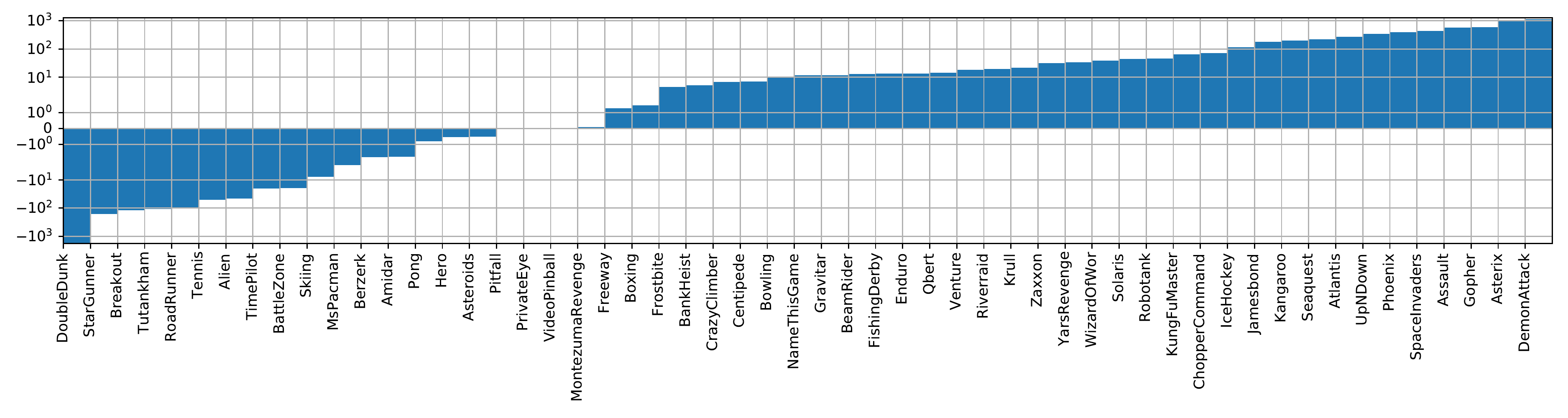}
    \caption{Improvement in normalized scores of TD-DQN over DDQN in 200M frames}
    \label{fig:improvement}
\end{figure}

\begin{figure}
    \centering
\end{figure}

The results clearly show that our proposed methods for guiding exploration, V-DQN and TD-DQN, both improve on the standard DDQN benchmark.  Although there are small differences in the ranking, both versions perform well in the same games, and underperform  the benchmark in a small set of games. The mean and median statistics do not reveal significant differences between V-DQN and TD-DQN.  Our intuition remains that V-DQN is likely to more sensitive to fluctuations and will allocate more exploration budget to damp them out.

\begin{table}
    \centering
    \begin{tabular}{lrrr}
    \hline
         & DDQN & V-DQN & TD-DQN \\
    \hline
        Median & 151\% & 164\% & 164\% \\
        Mean & 468\% & 547\% & 533\%\\
    \hline
    \end{tabular}
    \caption{Summary of normalized scores. See Table~\ref{tab:atari_normalized} in Appendix~\ref{appendix:results} for full results.}
    \label{tab:atari_summary}
\end{table}

\begin{figure}
    \centering
    \begin{minipage}{.48\textwidth}
        \includegraphics[width=\textwidth]{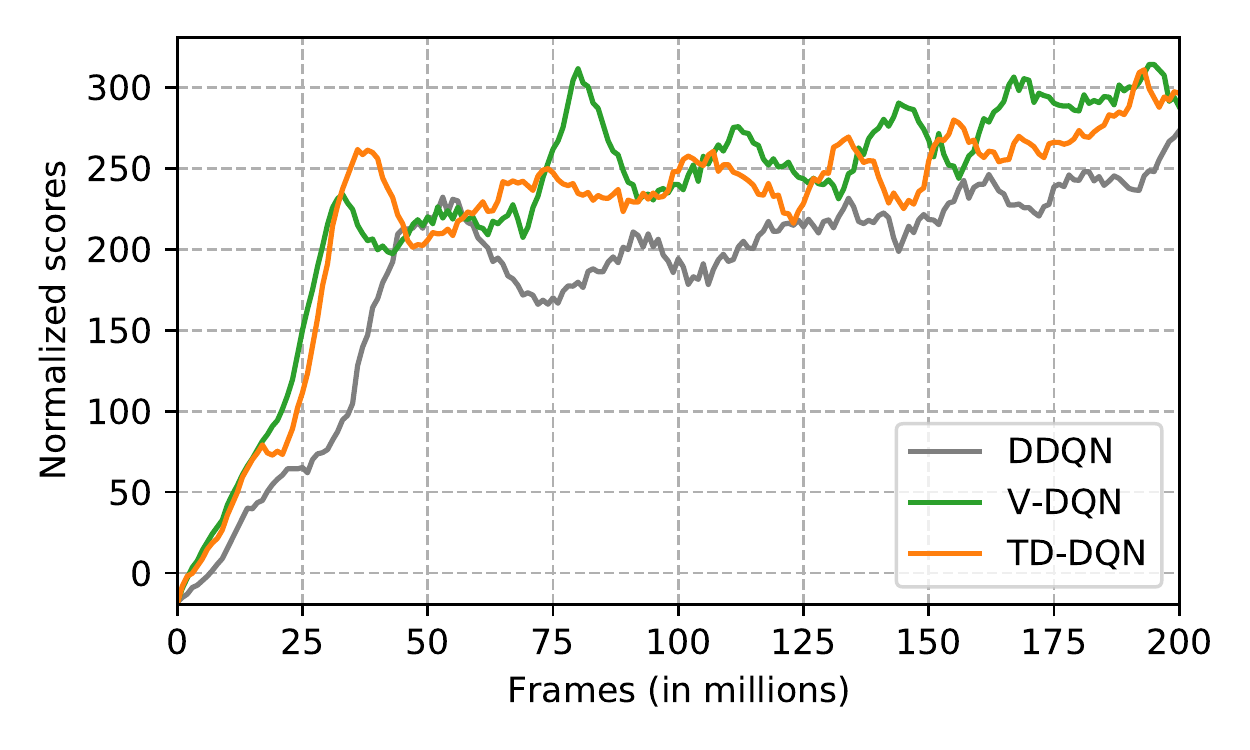}
        \subcaption{Mean normalized scores of all Atari games over 200M frames.}
    \end{minipage}
    \hfill
    \begin{minipage}{.48\textwidth}
        \includegraphics[width=\textwidth]{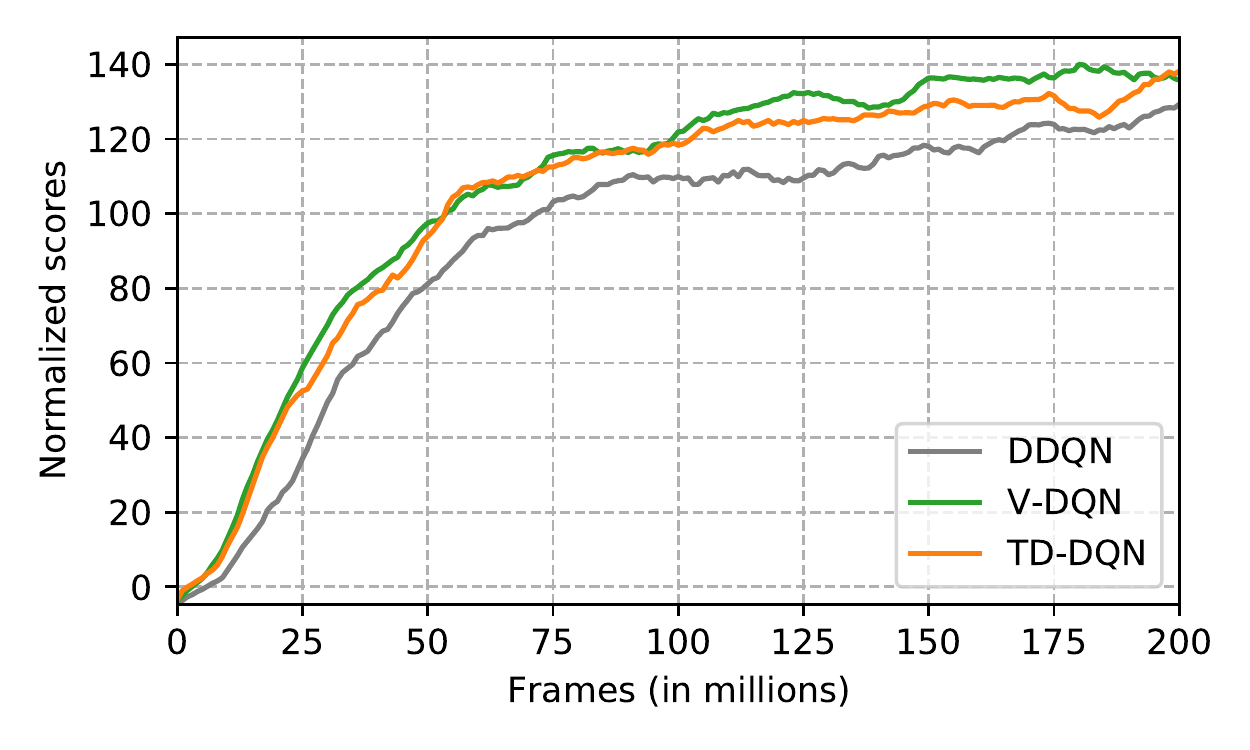}
        \subcaption{Median normalized scores of all Atari games over 200M frames.}
    \end{minipage}
    \caption{The mean and median of the normalized training curve over all 55 Atari games}
    \label{fig:atari_mean_median}
\end{figure}

\section{Conclusion}
\label{section:discussion}
This paper presents an analysis of the benefits and limitations of weighted variance for guiding exploration. Both weighted convergence and its close cousin, weighted temporal difference, can be used to quantify the rate of convergence of the return series for specific state-action pairs.  The return dynamics of value-based reinforcement learning is particularly susceptible to diverging as value improvements beget policy ones. This paper introduces a new two-stream network architecture to estimate both weighted variance/TD
errors; both our techniques (V-DQN and TD-DQN) outperform DDQN on the Atari game benchmark. We have identified two promising directions for future work in this area: 1) unifying exploration architectures to ameliorate the cold start issues and 2) adapting our exploration strategies to other deep reinforcement learning models.

\textbf{Addressing Cold Start with Unified Exploration} While our methods capture the divergence of return sequences, they suffer from the ``cold start'' problem.  It is unlikely that they will perform well for either empty or short sequences.  To address this, we propose two simple initialization methods for tabular configurations in this paper. This issue is somewhat alleviated by the generalization capacity inherent to function approximators like deep neural networks. However, larger state space where most of the states will never be visited still pose a problem. Our method can further benefit from unification with the other exploration methods. Count-based upper confidence bound (UCB) methods~\citep{Ostrovski2017count} have a stronger effect on balancing the visits among states in the early stage.  This effect decays gradually as visits increase. This characteristic makes it a natural complement for our sequence-based methods. Noisy DQN~\citep{Fortunato2018noisy} is another option that assigns greater randomness to less visited states. Our method focuses on promising actions whereas Noisy DQN chooses actions more randomly in those areas of the state space.  We ran some experiments on a rudimentary design in which the linear layers of Q-value stream were replaced with noisy ones; our preliminary results (not reported) show an improvement by hybridizing the two architectures.

\textbf{Beyond Q Learning} The key intuition behind our methods is that exploration should be guided with with a measure of historical variation.  Generally, the returns based on the estimated Q values of successive states are more consistent than those built on purely episodic experience. Therefore, it is natural to extend the idea of using return sequences to algorithms where state or action values are available, such as actor-critic methods. In policy gradient algorithms that strictly rely on observed reward, a history of individual action preferences is a good candidate for measuring variation.

\newpage
\bibliographystyle{iclr2021}
\bibliography{references.bib}

\appendix
\newpage
\section{Updating Formula in Variance Estimation}
\label{appendix:updating_formula}

Here we show that the weighted standard deviation in equation~\ref{eq:ve_variance} can be obtained by updating $\sigma$ with formula~\ref{eq:ve_update} in Variance Estimation. For simplicity, the inputs to all the following functions are hidden as they are the same state-action pair $(s, a)$.

While being updated with equation~\ref{eq:q_update}, we have
\begin{align}
    Q_n & = (1-\alpha)^n Q_0 + \sum_{i=1}^n \alpha (1 - \alpha)^{n-i} G_i \label{eq:qn} \\
    & = \alpha \sum_{i=1}^n (1 - \alpha)^{n-i} G_i \label{eq:q0}
\end{align}
The equal sign in formula~\ref{eq:q0} is established by setting $Q_0 = 0$.

If we expand $1 / \alpha$ as a power series
\begin{align}
    \frac{1}{\alpha} & = \frac{1 - (1 - \alpha)^n}{1 - (1 - \alpha)} \cdot \frac{1}{1 - (1 - \alpha)^n} \\
    & = \sum_{i=0}^{n-1} (1 - \alpha)^i \cdot \frac{1}{1 - (1 - \alpha)^n} \\
    & \approx \sum_{i=0}^{n-1} (1 - \alpha)^i \label{eq:power_series} \\
    & = \sum_{i=1}^n (1 - \alpha)^{n-i}
\end{align}
The approximate sign is used here to reflect that the second term of the multiplication approximates 1 as $n \xrightarrow{} \infty$ because $\alpha \in (0, 1]$. The value function can be re-written as an exponentially weighted sum of the return sequence by applying the above expansion of power series:
\begin{equation}
    Q_n \approx \frac{\sum_{i=1}^n (1 - \alpha)^{n-i} G_i}{\sum_{i=1}^n (1 - \alpha)^{n-i}}
    \label{eq:q_func}
\end{equation}

We denote $A_{n} = \sum_{i=1}^n (1 - \alpha)^{n - i}$, then
\begin{align}
    \sigma_{n+1}^2 & = \frac{\sum_{i=1}^{n+1}(1-\alpha)^{n+1-i}(G_i - Q_{n+1})^2}{A_{n+1}} \\
    & = \frac{\sum_{i=1}^{n}(1-\alpha)^{n+1-i}(G_i - Q_{n+1})^2}{A_{n+1}} + \frac{(G_{n+1} - Q_{n+1})^2} {A_{n+1}} \\
    & = (1 - \alpha) \cdot \frac{\sum_{i=1}^{n}(1-\alpha)^{n-i}(G_i - Q_{n+1})^2}{A_{n+1}} + \frac{(G_{n+1} - Q_{n+1})^2} {A_{n+1}} \\
    & = (1 - \alpha) \cdot \frac{\sum_{i=1}^{n}(1-\alpha)^{n-i}(G_i - Q_n + Q_n - Q_{n+1})^2}{A_{n+1}} + \frac{(G_{n+1} - Q_{n+1})^2} {A_{n+1}} \\
    &
    \begin{aligned}
        = & (1 - \alpha) \cdot \left[ \frac{\sum_{i=1}^{n}(1-\alpha)^{n-i}(G_i - Q_n)^2}{A_{n+1}} + \frac{\sum_{i=1}^{n}(1-\alpha)^{n-i}(Q_{n} - Q_{n+1})^2}{A_{n+1}} \right. \\
        & \left. + \frac{2 \sum_{i=1}^{n}(1-\alpha)^{n-i}(G_i - Q_n)(Q_n - Q_{n+1})}{A_{n+1}} \right] + \frac{(G_{n+1} - Q_{n+1})^2} {A_{n+1}} 
    \end{aligned} \\
    & \approx (1 - \alpha) \cdot \frac{A_n}{A_{n+1}} \cdot \left[ \sigma_n^2 + (Q_{n+1} - Q_n)^2 + 0 \right] + \frac{(G_{n+1} - Q_{n+1})^2} {A_{n+1}} \\
    & \approx (1 - \alpha) \left[ \sigma_n^2 + (Q_{n+1} - Q_{n})^2 \right] + \alpha (G_{n+1} - Q_{n+1})^2
\end{align}
where the first approximate sign comes from formula~\ref{eq:q_func} and the second one comes from formula~\ref{eq:power_series}.

Though the updating formula is biased as $n$ doesn't approach infinity in practice, the bias is negligible. Because all the above approximations are rooted in formula~\ref{eq:power_series} and $(1 - \alpha)^n$ converges to 0 rapidly as $n$ grows.

\newpage
\section{An Empirical Analysis}
\label{appendix:empirical_analysis}

\begin{figure}[H]
\centering
\begin{subfigure}[b]{\textwidth}
    \centering
    \includegraphics[width=\textwidth]{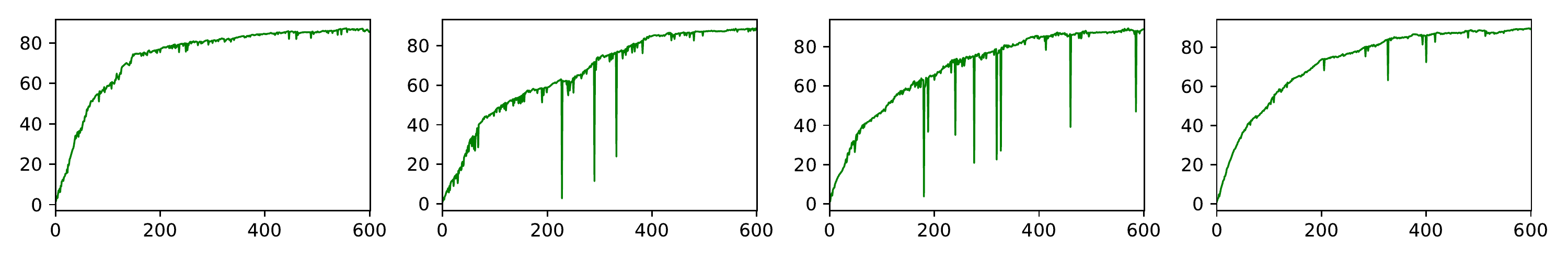}
    \caption{Return sequence}
    \label{fig:return_raw}
\end{subfigure}
\begin{subfigure}[b]{\textwidth}
    \centering
    \includegraphics[width=\textwidth]{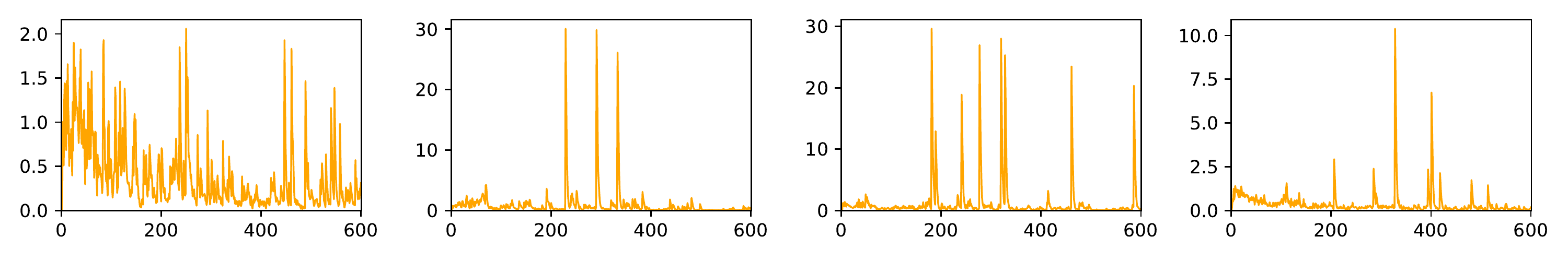}
    \caption{$\sigma$ value}
    \label{fig:sigma_raw}
\end{subfigure}
\caption{Raw return sequences (a) and corresponding $\sigma$ values (b) over visitations.}
\end{figure}

\begin{figure}[H]
\centering
\begin{subfigure}[b]{\textwidth}
    \centering
    \includegraphics[width=\textwidth]{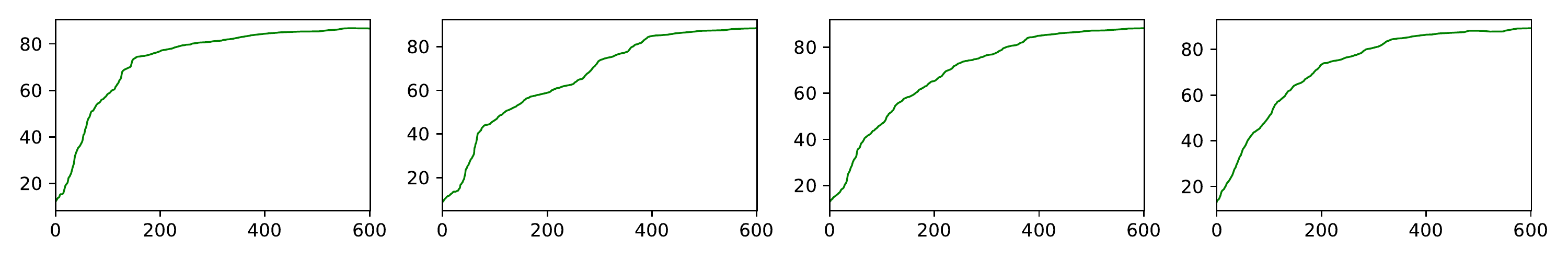}
    \caption{Return sequence}
    \label{fig:return_smoothed}
\end{subfigure}
\begin{subfigure}[b]{\textwidth}
    \centering
    \includegraphics[width=\textwidth]{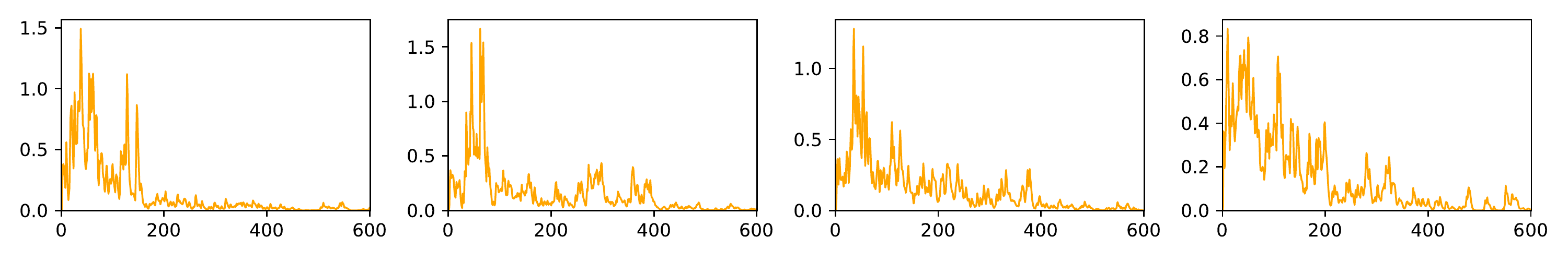}
    \caption{$\sigma$ value}
    \label{fig:sigma_smoothed}
\end{subfigure}
\caption{Smoothed return sequences (a) and corresponding $\sigma$ values (b) over visitations.}
\end{figure}

\begin{figure}[H]
\centering
\begin{subfigure}[b]{\textwidth}
    \centering
    \includegraphics[width=\textwidth]{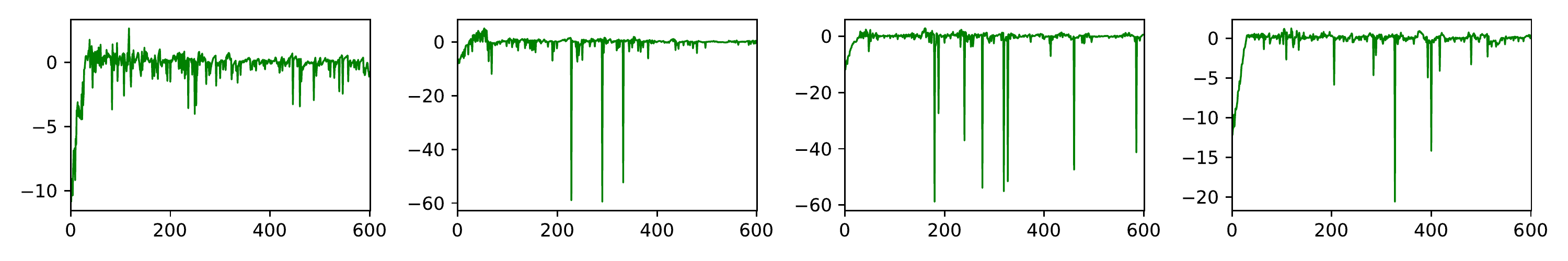}
    \caption{Return sequence}
    \label{fig:return_residual}
\end{subfigure}
\begin{subfigure}[b]{\textwidth}
    \centering
    \includegraphics[width=\textwidth]{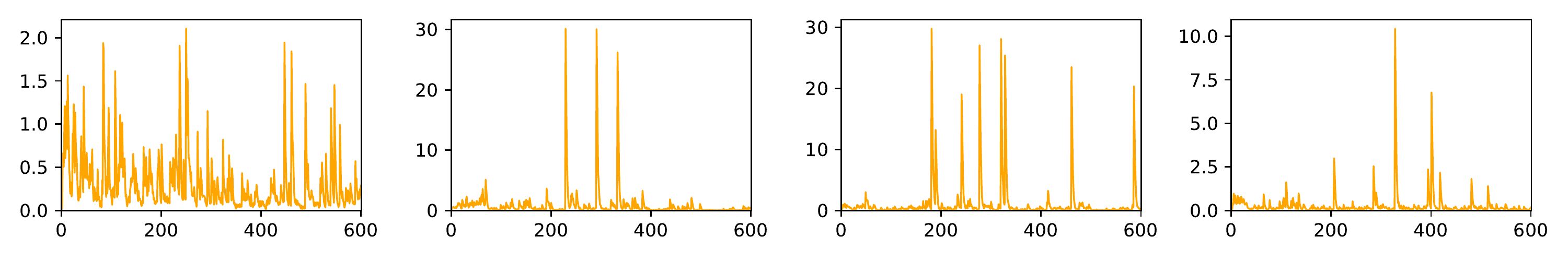}
    \caption{$\sigma$ value}
    \label{fig:sigma_residual}
\end{subfigure}
\caption{Residual return sequences (a) and corresponding $\sigma$ values (b) over visitations.}
\end{figure}

\cite{Sutton2018rl} prove that the policy and value functions monotonically improve over sweeps in policy iteration. In this simple form, return sequences for individual state-action pairs are monotonically increasing and usually converge fast. However, the same statement is not true when the policy is updated with Monte Carlo settings where the returns are random samples. In this situation, fluctuations in returns are inevitable due to the stochastic property; an incremental updating scheme is adopted to stabilize learning as well as avoid the excessive impact of malicious samples.

Here we illustrate the characteristics of the return sequences and analyze how variance guides exploration. The raw return sequences shown in Figure~\ref{fig:return_raw} are randomly sampled from frequently visited state-action pairs in the Cartpole balancing problem and truncated to be of the same length. To disentangle the impact of the convergence trend from transient fluctuations, smoothed versions of those sequences are extracted from the raw sequences (shown in Figure~\ref{fig:return_smoothed}), while the residuals are shown in Figure~\ref{fig:return_residual}. Then we apply our Variance Estimation method on each of the sequences independently and show the $\sigma$ values over visits.

Since the $\sigma$ values are directly integrated into the Q values used by our exploration policy (shown in Equation~\ref{eq:upper_bound}), a greater $\sigma$ value usually results in an increase in exploration budget. Meanwhile, the nature of weighted sum balances the importance of learned Q value and the history of its change which is captured by $\sigma$.

In Figure~\ref{fig:sigma_smoothed}, we observe that the $\sigma$ value is greater when the return sequence changes at a faster rate. As the sequence converges, the $\sigma$ value approaches zero. An interesting but unobvious observation is that the $\sigma$ value spikes when there is change in the convergence rate of the return sequence. Since variance is essentially a Euclidean distance metric, it is capable of capturing second-order information. 

Sequences in Figure~\ref{fig:return_residual} isolate the impacts of transient fluctuations from the overall trend. We observe that whenever there is an excessive fluctuation, the  $\sigma$ value spikes to a high magnitude to demand immediate exploration. Once the return goes back to its normal range, the $\sigma$ value decreases simultaneously. Those quick responses are useful since excessive fluctuations are harmful to the estimation of Q values. A timely investigation of exploration budget eliminates this negative impact before it propagates to more states. Meanwhile, frequent fluctuations beget an increase in $\sigma$ value and result in more exploration to determine its value. 

In conclusion, the exploration policy on our constructed upper bound effectively allocates exploration budget to accelerate convergence in important states as well as alleviate the impact of fluctuations.

\newpage
\section{Hyperparameters}
\label{appendix:hps}

\begin{table}[H]
    \centering
    \caption{Atari DQN Hyperparameters}
    \begin{tabular}{llp{75mm}}
        \hline
        Hyperparameter & Value & Description \\
        \hline
        $c_{\text{V-DQN}}$ & 0.1 & Weighting factor of $\sigma$-stream in exploration policy in V-DQN\\
        $c_{\text{TD-DQN}}$ & 0.1 & Weighting factor of $\sigma$-stream in exploration policy in TD-DQN\\
        mini-batch size & 32 & Size of mini-batch sample for gradient step \\
        replay buffer size & 1M & Maximum number of transitions stored in the replay buffer \\
        initial replay buffer size & 50K & Number of transitions stored in the replay buffer before optimization starts \\
        optimization frequency & 4 & Number of actions the agent takes between successive network optimization steps \\
        update frequency & 30000 & Number of steps between consecutive target updates \\
        $\epsilon_{\text{init}}$ & 1.00 & Initial exploration rate of $\epsilon$-greedy method \\
        $\epsilon_{\text{final}}$ & 0.01 & Final exploration rate of $\epsilon$-greedy method \\
        $N_{\epsilon}$ & 1M & Number of actions that the exploration rate of $\epsilon$-greedy method decays from initial value to final value \\
        $\alpha$ & 0.0000625 & Adam optimizer learning rate \\
        $\epsilon_{\text{ADAM}}$ & 0.00015 & Adam optimizer parameter \\
        evaluation frequency & 250K & Number of actions between successive evaluation runs \\
        evaluation length & 125K & Number of actions per evaluation run \\
        evaluation episode length & 27K & Maximum number of action in an episode in evaluation runs \\
        max no-op & 30 & Maximum number of no-op actions before the episode starts \\
    \end{tabular}
    \label{tab:hps}
\end{table}

\newpage
\section{Experimental Results on Atari Games}
\label{appendix:results}

\begin{figure}[H]
    \centering
    \includegraphics[width=\textwidth]{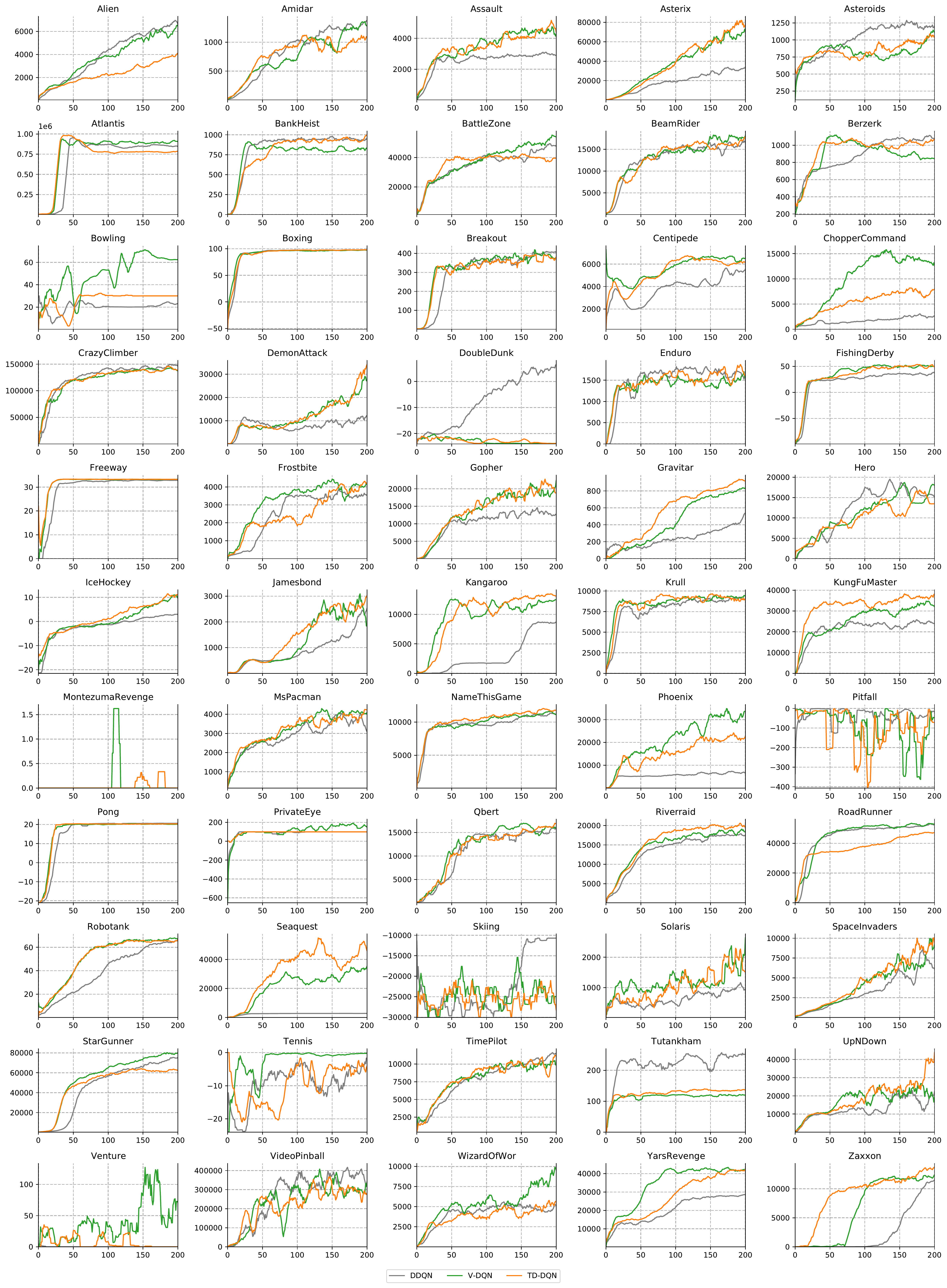}
    \caption{Training curve on Atari games from a single training run each. Episodes start with up to 30 no-op actions. Each data point is an average of episode rewards from 500K frames of evaluation runs, and smoothed over 10 data points.}
    \label{fig:atari_reward}
\end{figure}

\begin{table}[H]
    \centering
    \begin{tabular}{|l|l|ll|}
        \hline
        Game & DDQN & V-DQN & TD-DQN \\
        \hline
        Alien & 123\% & \textbf{130\%} & 76\% \\
        Amidar & \textbf{99\%} & 97\% & 97\% \\
        Assault & 764\% & \textbf{1659\%} & 1324\% \\
        Asterix & 537\% & 1280\% & \textbf{1490\%} \\
        Asteroids & \textbf{2\%} & 1\% & 1\% \\
        Atlantis & 7445\% & 7334\% & \textbf{7714\%} \\
        BankHeist & 165\% & 158\% & \textbf{170\%} \\
        BattleZone & 168\% & \textbf{199\%} & 149\% \\
        BeamRider & 143\% & 153\% & \textbf{156\%} \\
        Berzerk & \textbf{51\%} & 49\% & 49\% \\
        Bowling & 7\% & \textbf{52\%} & 18\% \\
        Boxing & 902\% & 903\% & \textbf{904\%} \\
        Breakout & 1764\% & \textbf{1851\%} & 1644\% \\
        Centipede & 57\% & 64\% & \textbf{64\%} \\
        ChopperCommand & 39\% & \textbf{237\%} & 111\% \\
        CrazyClimber & 639\% & \textbf{658\%} & 646\% \\
        DemonAttack & 563\% & 1415\% & \textbf{1700\%} \\
        DoubleDunk & \textbf{1568\%} & -145\% & -90\% \\
        Enduro & 267\% & \textbf{284\%} & 281\% \\
        FishingDerby & 151\% & \textbf{164\%} & 164\% \\
        Freeway & 130\% & \textbf{131\%} & 131\% \\
        Frostbite & 108\% & \textbf{116\%} & 112\% \\
        Gopher & 1071\% & \textbf{1689\%} & 1665\% \\
        Gravitar & 14\% & 22\% & \textbf{25\%} \\
        Hero & \textbf{83\%} & 81\% & 82\% \\
        IceHockey & 139\% & 227\% & \textbf{253\%} \\
        Jamesbond & 890\% & \textbf{1315\%} & 1071\% \\
        Kangaroo & 354\% & 529\% & \textbf{552\%} \\
        Krull & 909\% & 917\% & \textbf{931\%} \\
        KungFuMaster & 150\% & 199\% & \textbf{214\%} \\
        MontezumaRevenge & -1\% & \textbf{-0\%} & -1\% \\
        MsPacman & 32\% & \textbf{33\%} & 29\% \\
        NameThisGame & 207\% & 210\% & \textbf{218\%} \\
        Phoenix & 134\% & \textbf{852\%} & 531\% \\
        Pitfall & 5\% & \textbf{5\%} & 5\% \\
        Pong & \textbf{116\%} & 115\% & 115\% \\
        PrivateEye & -1\% & \textbf{-1\%} & -1\% \\
        Qbert & 146\% & 150\% & \textbf{160\%} \\
        Riverraid & 136\% & 152\% & \textbf{156\%} \\
        RoadRunner & 826\% & \textbf{827\%} & 724\% \\
        Robotank & 1006\% & \textbf{1071\%} & 1051\% \\
        Seaquest & 6\% & 122\% & \textbf{226\%} \\
        Skiing & \textbf{43\%} & 37\% & 35\% \\
        Solaris & -5\% & \textbf{50\%} & 40\% \\
        SpaceInvaders & 889\% & \textbf{1527\%} & 1318\% \\
        StarGunner & 922\% & \textbf{984\%} & 760\% \\
        Tennis & \textbf{197\%} & 147\% & 146\% \\
        TimePilot & \textbf{387\%} & 365\% & 366\% \\
        Tutankham & \textbf{215\%} & 108\% & 109\% \\
        UpNDown & 366\% & 516\% & \textbf{707\%} \\
        Venture & -1\% & \textbf{36\%} & 18\% \\
        VideoPinball & 425\% & \textbf{425\%} & 425\% \\
        WizardOfWor & 168\% & \textbf{309\%} & 206\% \\
        YarsRevenge & 61\% & \textbf{98\%} & 95\% \\
        Zaxxon & 151\% & 175\% & \textbf{183\%} \\
        \hline
    \end{tabular}
    \caption{Normalized scores.}
    \label{tab:atari_normalized}
\end{table}

\begin{table}[H]
    \centering
    \begin{tabular}{|l|ll|lll|}
        \hline
        Game & Random & Human & DDQN & V-DQN & TD-DQN \\
        \hline
        Alien & 128.3 & 6371.3 & 7807.3 & \textbf{8236.9} & 4895.4 \\
        Amidar & 11.8 & 1540.4 & \textbf{1521.6} & 1495.0 & 1494.0 \\
        Assault & 166.9 & 628.9 & 3697.5 & \textbf{7829.8} & 6284.9 \\
        Asterix & 164.5 & 7536.0 & 39782.0 & 94509.1 & \textbf{110013.6} \\
        Asteroids & 871.3 & 36517.3 & \textbf{1464.3} & 1317.6 & 1278.0 \\
        Atlantis & 13463.0 & 26575.0 & 989675.0 & 975100.0 & \textbf{1024975.0} \\
        BankHeist & 21.7 & 644.5 & 1050.1 & 1007.7 & \textbf{1082.3} \\
        BattleZone & 3560.0 & 33030.0 & 53153.8 & \textbf{62133.3} & 47460.0 \\
        BeamRider & 254.6 & 14961.0 & 21296.0 & 22765.7 & \textbf{23234.0} \\
        Berzerk & 196.1 & 2237.5 & \textbf{1228.1} & 1205.0 & 1190.7 \\
        Bowling & 35.2 & 146.5 & 42.7 & \textbf{93.6} & 54.7 \\
        Boxing & -1.5 & 9.6 & 98.6 & 98.7 & \textbf{98.8} \\
        Breakout & 1.6 & 27.9 & 465.5 & \textbf{488.3} & 434.1 \\
        Centipede & 1925.5 & 10321.9 & 6695.1 & 7271.9 & \textbf{7282.8} \\
        ChopperCommand & 644.0 & 8930.0 & 3900.0 & \textbf{20289.7} & 9835.0 \\
        CrazyClimber & 9337.0 & 32667.0 & 158346.2 & \textbf{162828.0} & 159952.0 \\
        DemonAttack & 208.3 & 3442.8 & 18418.2 & 45977.5 & \textbf{55200.6} \\
        DoubleDunk & -16.0 & -14.4 & \textbf{9.1} & -18.3 & -17.4 \\
        Enduro & -81.8 & 740.2 & 2113.5 & \textbf{2250.2} & 2225.0 \\
        FishingDerby & -77.1 & 5.1 & 46.6 & \textbf{57.7} & 57.7 \\
        Freeway & 0.1 & 25.6 & 33.2 & \textbf{33.6} & 33.6 \\
        Frostbite & 66.4 & 4202.8 & 4516.2 & \textbf{4883.8} & 4702.4 \\
        Gopher & 250.0 & 2311.0 & 22331.4 & \textbf{35061.4} & 34555.7 \\
        Gravitar & 245.5 & 3116.0 & 637.5 & 869.5 & \textbf{976.1} \\
        Hero & 1580.3 & 25839.4 & \textbf{21606.6} & 21244.0 & 21476.5 \\
        IceHockey & -9.7 & 0.5 & 4.5 & 13.5 & \textbf{16.1} \\
        Jamesbond & 33.5 & 368.5 & 3014.2 & \textbf{4438.8} & 3622.5 \\
        Kangaroo & 100.0 & 2739.0 & 9450.0 & 14057.6 & \textbf{14679.4} \\
        Krull & 1151.9 & 2109.1 & 9855.5 & 9930.0 & \textbf{10065.0} \\
        KungFuMaster & 304.0 & 20786.8 & 31070.7 & 40984.4 & \textbf{44177.4} \\
        MontezumaRevenge & 25.0 & 4182.0 & 0.0 & \textbf{9.1} & 3.3 \\
        MsPacman & 197.8 & 15375.0 & 5038.5 & \textbf{5246.7} & 4585.7 \\
        NameThisGame & 1747.8 & 6796.0 & 12181.1 & 12359.4 & \textbf{12774.7} \\
        Phoenix & 1134.4 & 6686.2 & 8568.6 & \textbf{48424.7} & 30623.9 \\
        Pitfall & -348.8 & 5998.9 & 0.0 & \textbf{0.0} & 0.0 \\
        Pong & -18.0 & 15.5 & \textbf{20.8} & 20.6 & 20.6 \\
        PrivateEye & 662.8 & 64169.1 & 100.0 & \textbf{200.0} & 100.0 \\
        Qbert & 183.0 & 12085.0 & 17551.4 & 18051.7 & \textbf{19250.0} \\
        Riverraid & 588.3 & 14382.2 & 19322.9 & 21545.0 & \textbf{22073.8} \\
        RoadRunner & 200.0 & 6878.0 & 55381.7 & \textbf{55437.1} & 48526.7 \\
        Robotank & 2.4 & 8.9 & 67.8 & \textbf{72.0} & 70.7 \\
        Seaquest & 215.5 & 40425.8 & 2789.8 & 49230.8 & \textbf{91277.1} \\
        Skiing & -15287.4 & -3686.6 & \textbf{-10314.1} & -10990.9 & -11215.2 \\
        Solaris & 2047.2 & 11032.6 & 1572.0 & \textbf{6497.6} & 5615.0 \\
        SpaceInvaders & 182.6 & 1464.9 & 11580.8 & \textbf{19757.7} & 17086.2 \\
        StarGunner & 697.0 & 9528.0 & 82076.7 & \textbf{87577.8} & 67768.6 \\
        Tennis & -21.4 & -6.7 & \textbf{7.5} & 0.2 & 0.0 \\
        TimePilot & 3273.0 & 5650.0 & \textbf{12460.7} & 11941.4 & 11975.0 \\
        Tutankham & 12.7 & 138.3 & \textbf{283.0} & 147.8 & 150.1 \\
        UpNDown & 707.2 & 9896.1 & 34346.4 & 48118.3 & \textbf{65673.9} \\
        Venture & 18.0 & 1039.0 & 9.6 & \textbf{382.9} & 197.1 \\
        VideoPinball & 20452.0 & 15641.1 & 584388.2 & \textbf{632013.8} & 631348.0 \\
        WizardOfWor & 804.0 & 4556.0 & 7115.1 & \textbf{12388.1} & 8551.7 \\
        YarsRevenge & 1476.9 & 47135.2 & 29332.9 & \textbf{46319.6} & 44894.4 \\
        Zaxxon & 475.0 & 8443.0 & 12488.4 & 14409.8 & \textbf{15028.3} \\
        \hline
    \end{tabular}
    \caption{Raw Scores.}
    \label{tab:atari_raw}
\end{table}

\end{document}